\documentclass{article} %

\usepackage[accepted]{arxiv}

\usepackage[utf8]{inputenc}
\usepackage{amsmath}
\usepackage{amssymb}
\usepackage{array}
\usepackage{xcolor} 
\definecolor{mydarkblue}{rgb}{0,0.08,0.45}
\usepackage[colorlinks,citecolor=mydarkblue,urlcolor=mydarkblue,linkcolor=mydarkblue,filecolor=mydarkblue]{hyperref}
\usepackage{url}
\usepackage{relsize}

\usepackage{multirow}
\usepackage{booktabs}
\usepackage{caption}
\usepackage{subcaption}
\usepackage{natbib}
\usepackage{graphicx}
\usepackage{makecell}
\usepackage{tabu}
\usepackage[hang,flushmargin]{footmisc} %

\setcitestyle{numbers,square}

\newcommand{\z}{\mathbf{z}}

\newcommand{\gaussk}{\mathcal{N}(\mathbf{\vmu}_k, \vSigma)}

\newcommand{\C}{\mathcal{C}}
\newcommand{\U}{\mathcal{U}}

\usepackage{amsmath,amsfonts,bm}

\def\eqref#1{equation~\ref{#1}}

\def\1{\bm{1}}

\def\vmu{{\bm{\mu}}}

\def\vx{{\bm{x}}}

\def\vz{{\bm{z}}}
\def\vmu{{\bm{\mu}}}
\def\vSigma{{\bm{\Sigma}}}

\DeclareMathAlphabet{\mathsfit}{\encodingdefault}{\sfdefault}{m}{sl}
\SetMathAlphabet{\mathsfit}{bold}{\encodingdefault}{\sfdefault}{bx}{n}

\usepackage{hhline}

\usepackage[textsize=scriptsize,textwidth=2.2cm]{todonotes}
\usepackage{etoolbox}
\newtoggle{showtodos}
 \usepackage[compact]{titlesec}
 \titlespacing{\section}{0pt}{1ex}{0ex}
 \titlespacing{\subsection}{0pt}{1ex}{0ex}

\begin{document}
\cfoot{\thepage}
\setlength{\footskip}{3em}

\twocolumn[
\icmltitle{Adversarial vulnerability of powerful near out-of-distribution detection}

%
%
%
%

%
%
%
%

%
%
%

\begin{icmlauthorlist}
\icmlauthor{Stanislav Fort}{stan}
\end{icmlauthorlist}

\icmlaffiliation{stan}{Stanford University}

\icmlcorrespondingauthor{Stanislav Fort}{sfort1@stanford.edu}

\icmlkeywords{Machine Learning, ICML}

\vskip 0.3in
]

%

%
%
%
%
%

%
\printAffiliationsAndNotice{} %

 \begin{abstract}
 There has been a significant progress in detecting out-of-distribution (OOD) inputs in neural networks recently, primarily due to the use of large models pretrained on large datasets, and an emerging use of multi-modality. 
 We show a severe adversarial vulnerability of even the strongest current OOD detection techniques. With a small, targeted perturbation to the input pixels, we can change the image assignment from an in-distribution to an out-distribution, and vice versa, easily. 
 In particular, we demonstrate severe adversarial vulnerability on the challenging near OOD CIFAR-100 vs CIFAR-10 task, as well as on the far OOD CIFAR-100 vs SVHN.
 We study the adversarial robustness of several post-processing techniques, including the simple baseline of Maximum of Softmax Probabilities (MSP), the Mahalanobis distance, and the newly proposed \textit{Relative} Mahalanobis distance. 
 By comparing the loss of OOD detection performance at various perturbation strengths, we demonstrate the beneficial effect of using ensembles of OOD detectors, and the use of the \textit{Relative} Mahalanobis distance over other post-processing methods. 
 In addition, we show that even strong zero-shot OOD detection using CLIP and multi-modality suffers from a severe lack of adversarial robustness as well.
 Our code is available on \href{https://github.com/stanislavfort/adversaries_to_OOD_detection}{GitHub}.
\end{abstract}

\section{Introduction}
The recent success of deep neural networks has led to their increasing deployment in high-stakes, safety critical applications such as health care \citep{dermatology,ren2019likelihood}, where models are required to be not only accurate but also robust to distribution shift. \citep{amodei2016concrete} Neural networks often assign high confidence to inputs that are misclassified, or even do not come from the distribution they were trained on at all \citep{guo2017calibration,lakshminarayanan2016simple}. Reliable out-of-distribution (OOD) detection remains a significant challenge.  

Improving OOD detection has seen progress by training generative models \citep{bishop1994novelty,nalisnick2019detecting, ren2019likelihood, morningstar2021density}, and modifying objective and loss functions \citep{zhang2020hybrid}. Exposure to a number of OOD samples during training has also lead to improvements \citep{hendrycks2018deep}.

Recently, large models (such as the Vision Transformer \citep{dosovitskiy2020image}) pre-trained on large datasets (such as ImageNet21k \citep{ridnik2021imagenet21k}) produced sufficiently high-quality image embeddings that allowed us to close the gap to human performance in many challenging near-OOD tasks in vision (such as distinguishing CIFAR-100 from CIFAR-10)\footnote{\url{https://paperswithcode.com/sota/out-of-distribution-detection-on-cifar-100-vs}}, as well as to make significant progress in genomics.~\citep{fort2021exploring}
\begin{figure}[h!]
    \centering
     \includegraphics[width=0.5\textwidth]{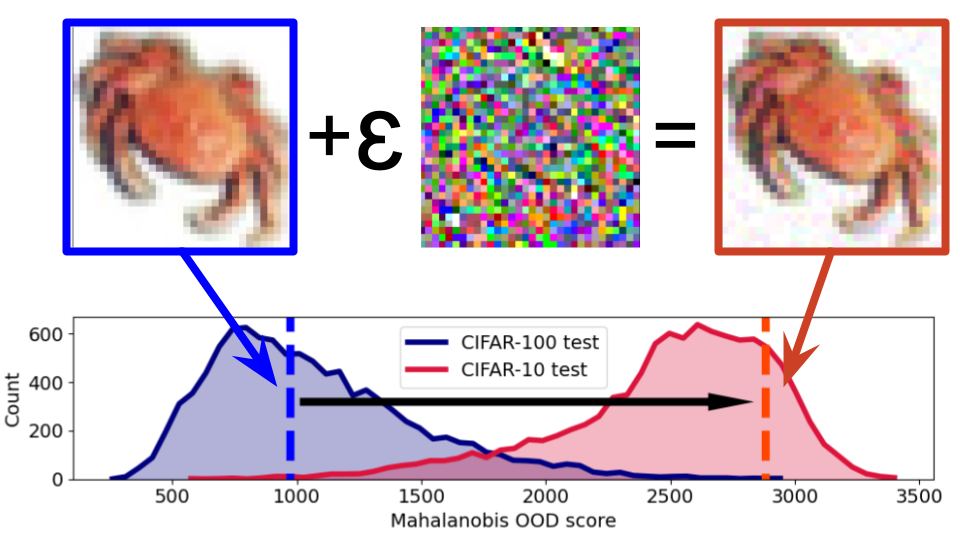}
    \caption{A small adversarial perturbation to the pixels of an in-distribution image (CIFAR-100) changes its out-of-distribution (OOD) score from $\approx$ 1,000 (around the mode of the in-distribution) to a confident out-distribution (CIFAR-10) region at $\approx$ 2,800 even for a state-of-the-art near-OOD detection method using a large ViT. $\varepsilon = 10^{-4}$ and the attack used is the Fast Gradient Sign Method applied to the Mahalanobis distance score for a ViT-L$_{16}$ as used in \citep{fort2021exploring}. The unperturbed CIFAR-100 $\to$ CIFAR-10 AUROC for this model is 97.98\%.}
    \label{fig:title_figure}
\end{figure}

A Mahalanobis distance (MD) based method \citep{lee2018simple} is a simple approach for post-processing embedding vectors coming from a neural network for OOD detection. Some of its common failure modes have been improved upon by the introduction of the \textit{Relative} Mahalanobis Distance (RMD) in \citep{ren2021simple}, generally improving performance while being more agnostic to hyperparameter choice.   

Mahalanobis distance based methods are good at detecting \emph{far} OOD samples -- for example CIFAR-10 vs SVHN, where the samples are distinct both in their surface-level style as well as in semantics. \emph{Near} OOD samples -- for example CIFAR-100 vs CIFAR-10, where samples are superficially similar and differ only in their semantic content -- have remained a challenge until the widespread use of large models and pre-training \citep{fort2021exploring}, and multi-modality (for example the use of CLIP \citep{radford2021learning} in \citep{fort2021exploring} for zero-shot class-name-only exposure OOD detection).  

The question of adversarial examples is usually framed in the classification setup, where an adversarial perturbation leads to a confident class change \citep{szegedy2013intriguing}. \citep{chen2020robust} show that OOD detection systems are also vulnerable to such attacks, and propose a robust training algorithm for counteracting it. \citep{2021ATOM} propose a training algorithm leading to more robust OOD detection as well.

\textbf{Key contributions:} We show empirically that currently even the most powerful and robust OOD detection systems based on large models and massive data are severely vulnerable to targeted adversarial attacks. We demonstrate that this is the case for different post-processing techniques, including the baseline Max of Softmax Probabilities (MSP), as well as the more advanced Mahalanobis distance. The zero-shot multi-modal approach using CLIP suffers from an even more acute vulnerability to such attacks. We show that working with lower resolution images increases OOD adversarial robustness. The largest positive effect we see comes from the use of ensembles of several OOD detectors, and the use of the \textit{Relative} Mahalanobis distance. We demonstrate that these two interventions can be successfully combined as well, making the detection system more adversarially robust as well as improving its OOD detection performance in general.

\section{Methods}
In this section, we describe how to get adversarial examples to OOD detection algorithms and briefly review the Mahalanobis distance and \textit{Relative} Mahalanobis distance methods. We also discuss the baseline Maximum of Softmax Probabilities, and the use of the multi-modal CLIP model for zero-shot OOD detection. We present a method for attacking ensembles of detectors we well.
\subsection{Generating adversarial examples to OOD score}
Given an out-of-distribution scoring function $\mathrm{score}(\vx)$ that maps an image $\vx$ into a floating point value characterizing its distance from the in-distribution, we can use its gradient with respect to the input,
\begin{equation}
    g(\vx) = \frac{\partial \mathrm{score}(\vx^\prime)}{\partial \vx^\prime}_{\vx^\prime = \vx} \, ,
\end{equation}
to gradually change the input $\vx$ to have either a higher or lower OOD score. This is exactly the same way adversarial examples, first described in \citep{szegedy2013intriguing}, are typically generated. Modifications exist that change the form of the perturbation, for example the \textit{Fast Gradient Sign Method} in \citep{goodfellow2014explaining} that uses $\mathrm{sign}(g(\vx))$ instead of $g(\vx)$ as the step direction. We will primarily be using that in this paper, as it is easy to use and works well out of the box.

Starting from an in-distribution image of a low score (confidently in-distribution), taking iterative steps
\begin{equation}
    \vx_{t+1} = \vx_{t} + \varepsilon g(\vx) \, ,
\end{equation}
where $\varepsilon$ is the learning rate, we move in the local direction of increasing OOD score. As shown in Figure~\ref{fig:title_figure}, a very small perturbation to an image of a \textit{crab} leads to a shift from the center of the in-distribution scores to the higher end of the out-distribution scores. This turns the image from a confidently and correctly in-distribution to a confidently out-distribution, as judged by a well-performing detection method from \citep{fort2021exploring}.
\subsection{Mahalanobis distance based OOD detection}
The Mahalanobis distance (MD)~\citep{lee2018simple} method and Relative Mahalanobis distance (RMD)~\citep{ren2021simple} method use intermediate features of a trained deep neural network. A frequent choice of the features are the pre-logits -- the output of the second to last layer of a network, just before the classification layer. Let us indicate these features as $\vz_i=f(\vx_i)$ for an input $\vx_i$.

For a $K$-class in-distribution dataset, both methods fit $K$ class-specific Gaussian distributions $\gaussk, k=1, 2, \dots, K$ to each of the $K$ in-distribution classes using their feature vectors $\vz_i$

We compute the class centroids (means) and covariance matrices as: 
   $ \mathbf{\vmu}_k = \frac{1}{N_k} \sum_{i:y_i=k} \vz_i$, for $k=1, \dots, K,$
   and $\vSigma = \frac{1}{N}\sum_{k=1}^K \sum_{i:y_i=k}\left( \vz_i-\mathbf{\vmu}_k \right)(\z_i-\mathbf{\vmu}_k)^T$.
Notice that the class means $\mathbf{\vmu}_k$ are independent for each class, while we use the same covariance matrix $\Sigma$ for all classes to avoid numerical issues due to under-fitting to the typically smaller than needed numbers of examples. 

For a test input $\vx'$ whose in- or out-distribution assignment is to be determined, we compute the Mahalanobis distances from the embedding vector of the test input $\vz'=f(\vx')$ to each of the $K$ in-distribution Gaussian distributions $\gaussk, k \in \{1,\dots,K\}$ given by $\text{MD}_k(\vz')$ we just computed. We take the minimum of the distances over all classes to be the uncertainty score $\U(\vx')$ characterizing how far from the in-distribution the input $\vx'$ is deemed to be. There the score can be seen as the extent to which the sample is OOD. The Mahalanobis distances are computed as 
\begin{align}
    \text{MD}_k(\vz') =& \left(\vz'-\mathbf{\vmu}_k\right)^T \vSigma^{-1} \left(\vz'-\mathbf{\vmu}_k \right), \\
     \mathrm{score}(\vx') =&  \U(\vx') 
     = - \min_{k} \{ \text{MD}_k(\vz') \}.
     \label{eq:maha}
\end{align}
This confidence score is used to distinguish the in-distribution and out-distribution samples from each other.
\subsection{Relative Mahalanobis Distance}
\label{sec:theory_ratio}
In \citep{ren2021simple} the \textit{Relative} Mahalanobis Distance is proposed which modifies Eq.~\ref{eq:maha} by subtracting a term to make it more robust to hyperparameter choice as well as generally better at OOD detection for near-OOD tasks in vision and genomics. The approach attempts to model the shape of the in-distribution and subtract its effects from the class-conditional distances. The RMD is defined as
\begin{align}
\text{RMD}_k(\vz') \nonumber = \text{MD}_k(\vz')  - \text{MD}_0(\vz') \, ,
\end{align}
where $\text{MD}_0(\vz')$ indicates the Mahalanobis distance to a Gaussian distribution fitted to the whole in-distribution dataset without regard to its label structure, as $\mathcal{N}(\mathbf{\vmu}_0, \vSigma_0)$, where $\vmu_0 = \frac{1}{N} \sum_{i=1}^N \z_i$ and $\vSigma_0 = \frac{1}{N}\sum_{i=1}^N \left( \vz_i-\vmu_0 \right)(\vz_i-\vmu_0)^T$. The goal is to model the background distribution. The resulting uncertainty score using RMD is then
\begin{align}
    \C^{\text{RMD}}(\vx') &= -\min_k \{ \text{RMD}_k(\vz') \}.
    \label{eq:ratio_score}
\end{align}
This can be extended to more powerful generative models fit (\citep{papamakarios2017masked, papamakarios2019normalizing}) to the class-specific and full-dataset approximations. \citep{ren2021simple}

\subsection{Maximum of Softmax Probabilities}
\label{sec:theory_MSP}
A solid baseline for OOD detection is provided by the simple approach of using the Maximum of Softmax Probabilities as the \textit{in}-distribution score. For a classification model $f(\vx) = \mathbf{p}$ that maps in input image $\vx$ to a vector of probabilities $\mathbf{p}$, the OOD score is $\mathrm{score}(\vx) = \mathrm{max}(f(\vx))$. For in-distribution images, for a well trained model the image will belong to one of the output classes that will likely be close to $1$ in the probabilities vector. For an OOD sample, the model will likely be confused and will not assign as high a probability to any of the classes. This provides the rational for using this method, which proved to be a good baseline given how simple its implementation is.

\subsection{Zero-shot multi-modal OOD detection using \textit{words} to specify distributions}
\label{sec:CLIP}
\citep{fort2021exploring} introduce a new kind of OOD detection scenario, where they use a multi-modal CLIP model \citep{radford2021learning}. CLIP produces a similarity score comparing the semantic content of an image and a text, as $\mathrm{logit}(\mathrm{I},\mathrm{T})$. By choosing two sets of words: in-words characterizing the semantic content of the in-distribution, and out-words, characterizing the semantic content of the out-distribution, for each image $I$ we can compute the in-logits for the in-words as $z^{\mathrm{in}}_i = \mathrm{CLIP}(I,\mathrm{inword}_i)$, and the out-logits for the out-words $z^{\mathrm{out}}_i = \mathrm{CLIP}(I,\mathrm{outword}_i)$. We construct the score the same way as in \citep{fort2021exploring} as
\begin{align}
\mathrm{score}(\vx) = &\mathrm{max}(\{\mathrm{CLIP}(\vx,\mathrm{inword}_i\}_i) \\ &-\mathrm{max}(\{\mathrm{CLIP}(\vx,\mathrm{outword}_i\}_i)  \, .
\end{align}
We can modify this score the same way we do for the Mahalanobis distance or Relative Mahalanobis distance using a gradient step with respect to the image.

\subsection{Ensembling OOD detectors}
\label{sec:ensemble}
A simple way to improve the OOD detection capabilities of several OOD detectors is to ensemble them. For example, this is used in \citep{fort2021exploring} to reach the current state-of-the-art performance on the near OOD CIFAR-100 $\to$ CIFAR-10 task. The simplest technique we can use is to generate the OOD score for a particular image $\vx$ for each of the models $\mathrm{score}_i(\vx)$, and compute their average 
\begin{equation}
\mathrm{score}_\mathrm{ensemble} (\vx) = \frac{1}{N} \sum_{i=1}^N \mathrm{score}_i(\vx) \, .
\end{equation}
The likely reason for why ensembling of the OOD predicted scores over several models works better than the models individually is similar to the reason for why deep ensembles work in general \citep{lakshminarayanan2016simple}. A loss landscape approach to that is discussed in \citep{fort2019deep}.

\subsection{Attacks on model ensembles}
Attacking an ensemble of OOD detectors, as discussed in Section~\ref{sec:ensemble}, is the same as attacking a single model. The only difference is that we replace the single model OOD scoring function $\mathrm{score}(\vx)$ with the ensemble scoring function $\mathrm{score}_\mathrm{ensemble}(\vx)$.

\section{Experiments and Results}

We studied the adversarial robustness of the currently best performing methods on the near-OOD task of distinguishing CIFAR-100 (in-distribution) from CIFAR-10 (out-distribution).\footnote{\url{https://paperswithcode.com/sota/out-of-distribution-detection-on-cifar-100-vs}} The best performing approach is an ensemble of pre-trained Vision Transformers finetuned on CIFAR-100 with the Mahalanobis distance post-processing method applied on top of their embeddings. This reaches an AUROC of 97.98\% \citep{fort2021exploring}, as compared to a human benchmark of AUROC $\approx$96.0\%. The best approach not using an ensemble of detectors differs in using a single ViT only.

We chose the pre-trained and finetuned ViT-L\_{16}\footnote{\url{https://github.com/google-research/vision_transformer}} to develop OOD adversarial attacks to. Its default resolution is $384\times384$ and we used the standard \emph{tf.image.resize} to upsample the $32\times32$ CIFAR images to it, as done in the standard ViT preprocessing pipeline.

\subsection{Attacks on CIFAR-100 vs CIFAR-10 for different post-processing techniques}

\begin{figure}[h!]
    \centering
     \includegraphics[width=0.40\textwidth]{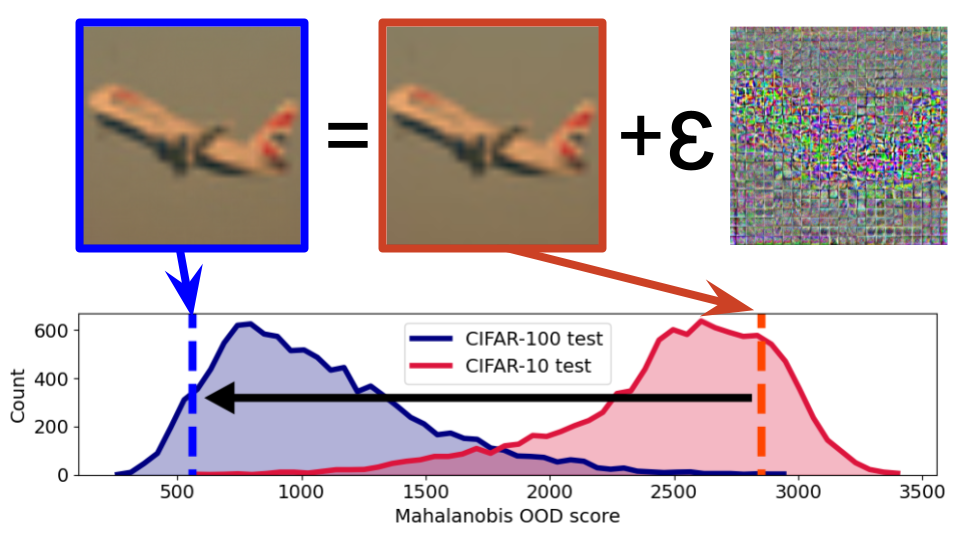}
    \caption{A small adversarial perturbation to the pixels of an out-distribution image (CIFAR-10) changes its out-of-distribution (OOD) score from $\approx$ 2,800 (around the mode of the out-distribution) to a confident in-distribution (CIFAR-100) region at $\approx$ 600 even for a state-of-the-art near-OOD detection method using a large ViT. $\varepsilon = 10^{-4}$ and the attack used is the Fast Gradient Sign Method applied to the Mahalanobis distance score for a ViT-L$_{16}$ as used in \citep{fort2021exploring}. The unperturbed CIFAR-100 $\to$ CIFAR-10 AUROC for this model is 97.98\%.}
    \label{fig:example_Maha}
\end{figure}

\paragraph{Mahalanobis distance}
We focused on the challenging near-OOD CIFAR-100 vs CIFAR-10 task. Figure~\ref{fig:example_Maha} shows an image of an airplane (CIFAR-10, out-distribution) being adversarially modified using the Fast Gradient Sign Method to read as a confident in-distribution image do the ViT based Mahalanobis distance OOD detector. The figure also shows the shift of the OOD score against the histograms of the in- and out-distribution test set images. This is similar to Figure~\ref{fig:title_figure}, where the direction of change was from the in-distribution to the out-distribution.
A small change in the pixel values of the input image resulted in a large change of the OOD score assigned.

\begin{figure}[h!]
    \centering
    \begin{subfigure}[b]{0.44\columnwidth}
         \centering
         \includegraphics[width=\textwidth]{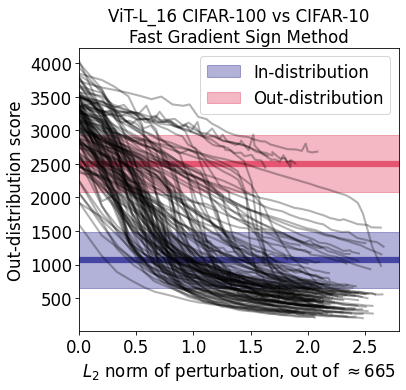}
         \caption{}
         \label{fig:bands_L2_maha}
     \end{subfigure} 
     \vspace{-0.5em}
    \begin{subfigure}[b]{0.44\columnwidth}
        \centering
        \includegraphics[width=\columnwidth]{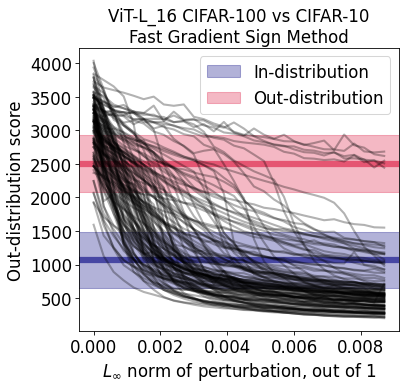}
        \caption{}
        \label{fig:bands_Linfty_maha}
     \end{subfigure} 
     \vspace{-0.6em}
    \caption{Changing the out-distribution score for a set of 128 CIFAR-10 test images (out-distribution) by applying the Fast Gradient Sign Method to the Mahalanobis OOD score based on a ViT-L\_16. (a) shows the score as a function of the $L_2$ norm of the image perturbation, while (b) shows the $L_\infty$ norm.}
\end{figure}

Applying the same procedure to 128 test images, we were able to generate a set of perturbed out-distribution images that read as confidently in-distribution to the detector, as shown in Figure~\ref{fig:bands_L2_maha} as a function of the $L_2$ norm of the image perturbation and in Figure~\ref{fig:bands_Linfty_maha} as a function of the $L_\infty$ norm.

\begin{figure}[h!]
    \centering
     \includegraphics[width=0.40\textwidth]{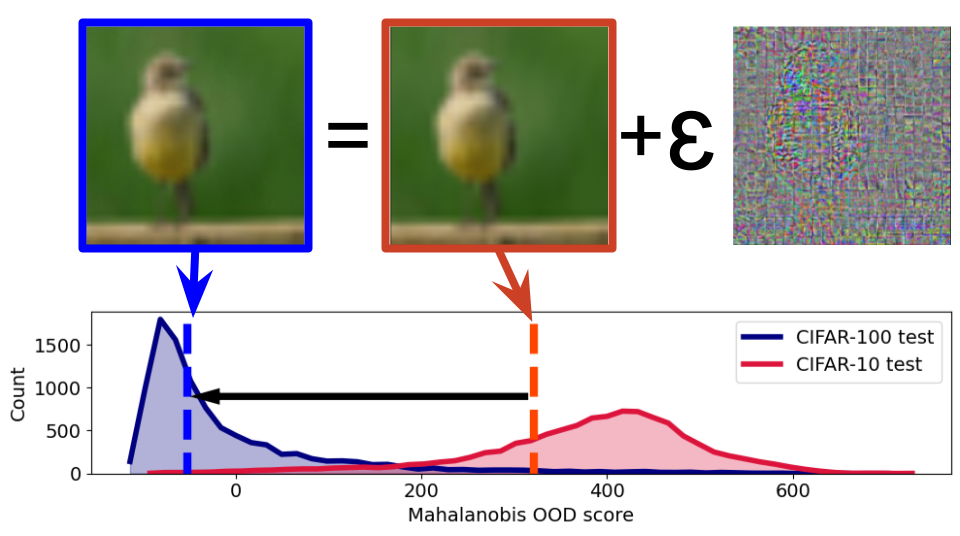}
    \caption{A small adversarial perturbation to the pixels of an out-distribution image (CIFAR-10) changes its out-of-distribution (OOD) score from $\approx$ 300 (around the mode of the out-distribution) to a confident in-distribution (CIFAR-100) region at $\approx$ -50 even for a state-of-the-art near-OOD detection method using a large ViT. $\varepsilon = 10^{-4}$ and the attack used is the Fast Gradient Sign Method applied to the \textit{Relative} Mahalanobis distance score for a ViT-L$_{16}$ as used in \citep{fort2021exploring}. The unperturbed CIFAR-100 $\to$ CIFAR-10 AUROC for this model is 97.11\%.}
    \label{fig:example_ratio}
\end{figure}

\begin{figure}[h!]
    \centering
    \begin{subfigure}[b]{0.44\columnwidth}
         \centering
         \includegraphics[width=\textwidth]{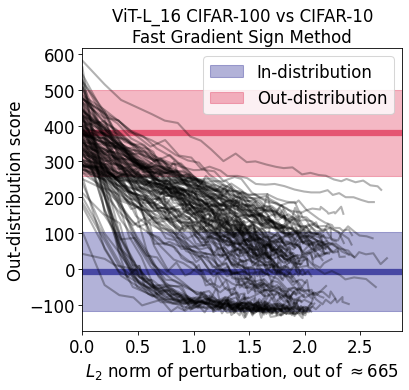}
         \caption{}
         \label{fig:bands_L2_ratio}
     \end{subfigure} 
     \vspace{-0.5em}
    \begin{subfigure}[b]{0.44\columnwidth}
        \centering
        \includegraphics[width=\columnwidth]{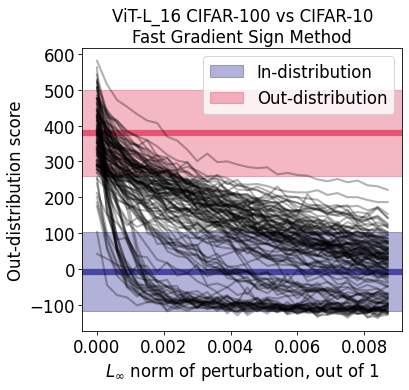}
        \caption{}
        \label{fig:bands_Linfty_ratio}
     \end{subfigure} 
     \vspace{-0.6em}
    \caption{Changing the out-distribution score for a set of 128 CIFAR-10 test images (out-distribution) by applying the Fast Gradient Sign Method to the \textit{Relative} Mahalanobis OOD score based on a ViT-L\_16. (a) shows the score as a function of the $L_2$ norm of the image perturbation, while (b) shows the $L_\infty$ norm.}
\end{figure}

\begin{figure}[h!]
    \centering
    \begin{subfigure}[b]{0.44\columnwidth}
         \centering
         \includegraphics[width=\textwidth]{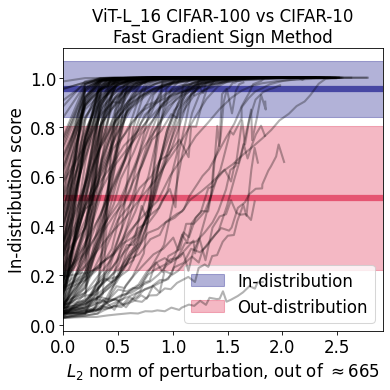}
         \caption{}
         \label{fig:bands_L2_MSP}
     \end{subfigure} 
     \vspace{-0.5em}
    \begin{subfigure}[b]{0.44\columnwidth}
        \centering
        \includegraphics[width=\columnwidth]{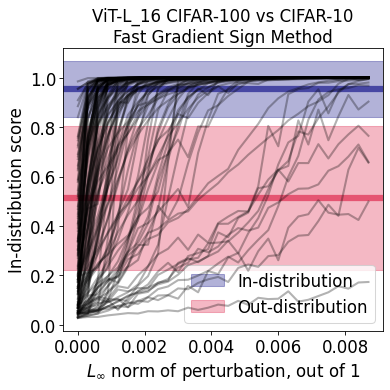}
        \caption{}
        \label{fig:bands_Linfty_MSP}
     \end{subfigure} 
     \vspace{-0.6em}
    \caption{Changing the out-distribution score for a set of 128 CIFAR-10 test images (out-distribution) by applying the Fast Gradient Sign Method to the Maximum of Softmax Probabilities (MSP) score based on a ViT-L\_16. (a) shows the score as a function of the $L_2$ norm of the image perturbation, while (b) shows the $L_\infty$ norm.}
    \vspace{-0.3em}
\end{figure}

\paragraph{Relative Mahalanobis distance}
Using the proposed Relative Mahalanobis distance \citep{ren2021simple}, that we discuss in Section~\ref{sec:theory_ratio}, as an OOD score, we show an equivalent effect of a small adversarial perturbation on the OOD score in Figure~\ref{fig:example_ratio}.
Applying this attack to 128 out-distribution images and their gradual score change with the $L_2$ and $L_\infty$ norms of the perturbation are shown in Figure~\ref{fig:bands_L2_ratio} and Figure~\ref{fig:bands_Linfty_ratio} respectively.
\paragraph{Maximum of Softmax Probabilities}
We used the Maximum of Softmax Probabilities (MSP) as a baseline post-processing method for OOD detection, as discussed in Section~\ref{sec:theory_MSP}. Figure~\ref{fig:bands_L2_MSP} and Figure~\ref{fig:bands_Linfty_MSP} show the change in the score of 128 out-distribution test images as a function of the $L_2$ and $L_\infty$ norms of the image perturbation.

\begin{figure}[h!]
    \centering
    \begin{subfigure}[b]{0.49\columnwidth}
         \centering
         \includegraphics[width=\textwidth]{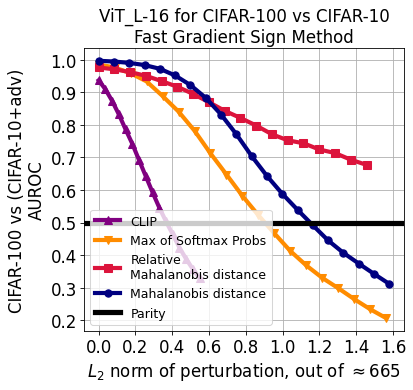}
         \caption{}
         \label{fig:auroc_L2_ViT-L_16}
     \end{subfigure} 
    \begin{subfigure}[b]{0.49\columnwidth}
        \centering
        \includegraphics[width=\columnwidth]{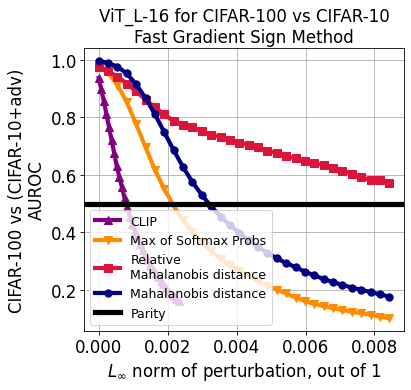}
        \caption{}
        \label{fig:auroc_Linfty_ViT-L_16}
     \end{subfigure} 
    \caption{AUROC of CIFAR-100 vs CIFAR-10 where the out-distribution CIFAR-10 is represented by 128 adversarially perturbed images to lower the Mahalanobis distance OOD score (blue) or Relative Mahalanobis distance score (red), and as a baseline the Maximum of Softmax Probabilities (orange). We add the CLIP zero-shot OOD detection as comparison in purple. (a) shows the perturbation strength measured by its $L_2$ norm, and (b) by its $L_\infty$ norm. Details in Table~\ref{tab:main_table} and Table~\ref{tab:main_table_CLIP_only}.}
\end{figure}

\paragraph{Robustness comparison}
\begin{table}[ht]
\begin{center}	
\caption{The loss of AUROC on the near OOD CIFAR-100 vs CIFAR-10 task for several OOD detection approaches. The strength of the attack is fixed by the $L_\infty$ norm of the adversarial perturbation at 1/255. All approaches use the pretrained and finetuned ViT-L$_{16}$ to generate probability outputs and embeddings. The baseline of using the Max of Softmax Probabilities (MSP) is the least robust, followed by the Standard Mahalanobis Distance. The newly proposed Relative Mahalanobis Distance has the highest adversarial robustness by a significant margin.}
\vspace{0.5em}
\begin{tabular}{ c|c|c|c }

\makecell{Post-process\\method} & \makecell{AUROC\\before} & \makecell{AUROC\\$\ell_{\infty}$\\1/255} & \makecell{$\Delta$\\AUROC} \\
\hline
\makecell{Max of\\Softmax Probs} & 94.28\% & 27.48\% & -66.8\% \\
\hline
Maha & 97.98\% & 41.33\% & -56.65\% \\
\hline
Relative Maha & 97.11\% & 71.84\% & \textbf{-25.27}\% \\
\hline
\end{tabular}
\label{tab:main_table}
\end{center}
\end{table}
The stronger the adversarial attack, the more we can change the out-distribution samples in order for them to be perceived as in-distribution by the detection system. Table~\ref{tab:main_table} summarizes the loss of the AUROC on the CIFAR-100 vs CIFAR-10 task for the standard Mahahalanobis distance, the Relative Mahalanobis distance, and the Maximum of Softmax Probabilities (comparison baseline). Figure~\ref{fig:auroc_L2_ViT-L_16} and Figure~\ref{fig:auroc_Linfty_ViT-L_16} show the loss of AUROC as a function of the perturbation strength measured by their $L_2$ and $L_\infty$ norms. The results in Table~\ref{tab:main_table} can be read off from Figure~\ref{fig:auroc_Linfty_ViT-L_16} by looking at $L_
\infty = 1/255$.

The way we turned Figures~\ref{fig:bands_Linfty_maha}, \ref{fig:bands_Linfty_ratio} and \ref{fig:bands_Linfty_MSP} into the summary in Figure~\ref{fig:auroc_Linfty_ViT-L_16} was as follows. Each image is adversarially modified in $T$ steps. Its OOD score and $L_\infty$ perturbation norm change as a function of $T$. We used a piece-wise linear interpolation to make an image-specific function $\mathrm{score}(L_\infty)$. Then, when making Figure~\ref{fig:auroc_Linfty_ViT-L_16}, we sampled the $L_\infty$ perturbation norms we wanted to explore, and for each computed the interpolated OOD score for each of the 128 images based on their individual linear interpolations. The resulting distribution of scores was then compared to the scores of the in-distribution test set to obtain the AUROC. For the $L_2$ norm in Figure~\ref{fig:auroc_L2_ViT-L_16} the process was analogous, swapping $L_\infty$ for $L_2$ everywhere.

To compare the robustness of the standard Mahalanobis distance and the \textit{Relative} Mahalanobis distance to OOD adversarial attacks, we used the Fast Gradient Sign Method of finding the adversary, with a learning rate of $3\times10^{-4}$ (arbitrarily chosen), and ran it for 30 steps on 128 test set images of CIFAR-10 (the out-distribution). We ran the attack against both the Mahalanobis distance score as well as the Relative Mahalanobis distance score. For each of the images, we measured its score, its $L_2$ distance from the unperturbed image (out of $\sqrt{384\times384\times3} \approx 665$ for fully saturated pixels in the $[0,1]$ range), and its $L_\infty$ distance from the unperturbed image (out of 1). 
For both the $L_2$ and $L_\infty$ perturbation strength norms, the Relative Mahalanobis distance is significantly more robust to OOD adversarial perturbations, retaining a higher AUROC on the near-OOD CIFAR-100 vs CIFAR-10 task at a given strength of perturbation. This is in line with the observation of higher stability of the relative distance method \citep{ren2021simple}. The baseline method of Maximum of Softmax Probabilities (in orange) performs the worst, losing AUROC the fastest with perturbation strength.

\subsection{Zero-shot OOD using CLIP}
\begin{table}[ht]
\begin{center}	
\caption{The loss of AUROC on the near OOD CIFAR-100 vs CIFAR-10 task for the CLIP zero-shot method using class names at $L_\infty$ of 1/255 perturbation strength. CLIP is by far the least robust technique we studied in this paper.}
\vspace{0.5em}
\begin{tabular}{ c|c|c|c }

\makecell{Post-process\\method} & \makecell{AUROC\\before} & \makecell{AUROC\\$\ell_{\infty}$\\1/255} & \makecell{$\Delta$\\AUROC} \\
\hline
CLIP & 94.68\% & $<$10\% & $\mathrm{a\,lot}$ \\
\hline
\end{tabular}
\label{tab:main_table_CLIP_only}
\end{center}
\end{table}
We use the zero-shot OOD detection setup using the multi-modal CLIP model described in Section~\ref{sec:CLIP} and introduced in \citep{fort2021exploring}. In Figures~\ref{fig:auroc_L2_ViT-L_16} and \ref{fig:auroc_Linfty_ViT-L_16} we show that its adversarial robustness is lower than for other methods, including the baseline Max of Softmax Probabilities (MSP). In Table~\ref{tab:main_table_CLIP_only} we show the underlying numbers in detail. Despite its versatility and power, CLIP does not perform very well when under a targeted adversarial attack to its OOD capabilities, underperforming even a simple post-processing baseline (albeit with very strong embeddings from a large, pretrained ViT).

\begin{figure}[h!]
    \centering
    \begin{subfigure}[b]{0.44\columnwidth}
         \centering
         \includegraphics[width=\textwidth]{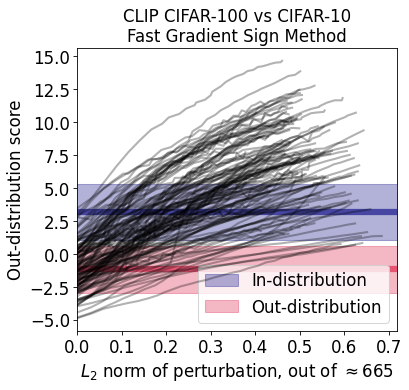}
         \caption{}
         \label{fig:bands_L2_CLIP}
     \end{subfigure} 
     \vspace{-0.5em}
    \begin{subfigure}[b]{0.44\columnwidth}
        \centering
        \includegraphics[width=\columnwidth]{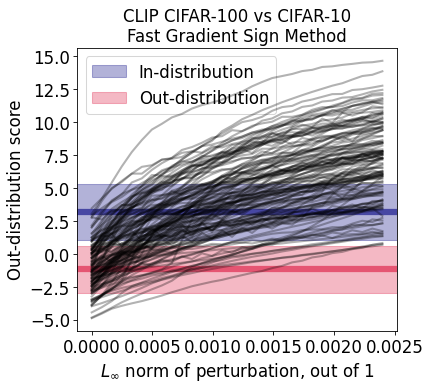}
        \caption{}
        \label{fig:bands_Linfty_CLIP}
     \end{subfigure} 
     \vspace{-0.3em}
    \caption{Changing the out-distribution score for a set of 128 CIFAR-10 test images (out-distribution) by applying the Fast Gradient Sign Method to the CLIP model. (a) shows the score as a function of the $L_2$ norm of the image perturbation, while (b) shows the $L_\infty$ norm.}
\end{figure}

The change in the OOD score for 128 test set images from the out-distribution under an adversarial attack against the CLIP-based detector is shown in Figure~\ref{fig:bands_L2_CLIP} for the $L_2$ norm of the perturbation strength and in Figure~\ref{fig:bands_Linfty_CLIP} for the $L_\infty$ norm.

\vspace{-0.3em}
\subsection{Model ensembles}
We studied ensembles of OOD detectors, as discussed in Section~\ref{sec:ensemble}. We used the standard setup using the Fast Gradient Sign Method (keeping only the sign of each element of the gradient), learning rate of $3\times10^{-4}$ (arbitrarily chosen) and ran it for 30 steps on 128 test set images of CIFAR-10 (the out-distribution). We identified two well performing models finetuned on CIFAR-100 (training set), the ViT-L$_{16}$ and R50+ViT-L$_{32}$, both with input resolution of $224\times224\times3$.

We found that OOD model ensembling: 1) improves OOD detection AUROC, 2) makes it more robust to adversarial attacks, and 3) its benefit combines well with the benefit of using the Relative Mahalanobis distance.

\begin{figure}[h!]
    \centering
     \includegraphics[width=0.20\textwidth]{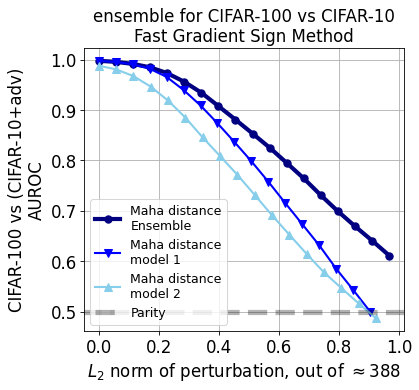}
    \includegraphics[width=0.20\textwidth]{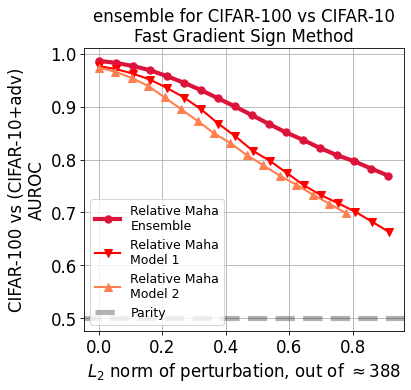}
    \caption{AUROC of CIFAR-100 vs CIFAR-10 where the out-distribution CIFAR-10 is represented by 128 adversarially perturbed images to lower the Mahalanobis distance OOD score (\textbf{left panel}, blue) and Relative Mahalanobis distnace score (\textbf{right panel}, red). We show the perturbation strength measured by its $L_2$ norm. The model ensemble (darker lines) is more robust to adversarial perturbations both for the standard and relative distance post-processing.}
    \label{fig:ensemble_L2}
\end{figure}

\begin{figure}[h!]
    \centering
     \includegraphics[width=0.23\textwidth]{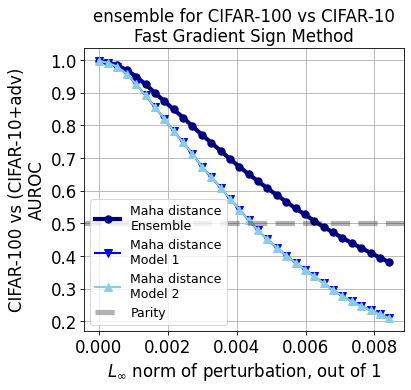}
    \includegraphics[width=0.23\textwidth]{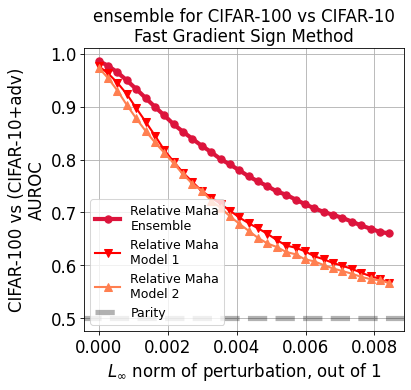}
    \caption{AUROC of CIFAR-100 vs CIFAR-10 where the out-distribution CIFAR-10 is represented by 128 adversarially perturbed images to lower the Mahalanobis distance OOD score (\textbf{left panel}, blue) and Relative Mahalanobis distnace score (\textbf{right panel}, red). We show the perturbation strength measured by its $L_\infty$ norm. The model ensemble (darker lines) is more robust to adversarial perturbations both for the standard and relative distance post-processing.}
    \label{fig:ensemble_Linfty}
\end{figure}

We show the detailed results in Table~\ref{tab:ensemble} and in Figure~\ref{fig:ensemble_L2} and Figure~\ref{fig:ensemble_Linfty}.

We look at the performance of two models individually, and perform adversarial attacks on their OOD score. We record the drop in AUROC for distinguishing the unperturbed CIFAR-100 from the adversarially perturbed CIFAR-10 at the perturbation level $\ell_\infty = 1/255$. We do the same for the ensemble of the two models. Ensembles suffer from a smaller drop in AUROC at a given perturbation level. Its benefit can be combined with the large robustness benefit of the \textit{Relative} Mahalanobis distance. 
\begin{table}[ht]
\begin{center}	
\caption{The benefit of OOD detector ensembling for adversarial robustness. The results are shown for the near OOD CIFAR-100 $\to$ CIFAR-10 task. We evaluate two separate models, and their ensemble, each for using the Mahalanobis distance and the Relative Mahalanobis distance post-processing. Using an ensemble increases adversarial robustness, and can be combined to increase its benefit with the Relative Mahalanobis distance.}
\vspace{-0.5em}
\begin{tabular}{ c|c|c|c|c }

Model & \makecell{Post-\\process\\method} & \makecell{AUROC\\before} & \makecell{AUROC\\$\ell_{\infty}$ 1/255} & \makecell{$\Delta$\\AUROC} \\
\hline

ViT L$_{16}$ & Maha & 97.72\% & 56.14\% & -41.58\% \\
R50+L$_{32}$ & Maha & 96.95\% & 54.94\% & -42.01\% \\
Ensemble & Maha & 97.91\% & 68.67\% & \textbf{-29.24\%} \\
\hline

ViT L$_{16}$ & Relative & 96.92\% & 69.82\% & -27.10\% \\
R50+L$_{32}$ & Relative & 97.09\% & 68.53\% & -28.56\% \\
Ensemble & Relative & 97.69\% & 78.64\% & \underline{\textbf{-9.05\%}} \\

\hline

\end{tabular}
\label{tab:ensemble}
\vspace{-0.35em}
\end{center}
\end{table}
The AUROC as a function of the perturbation strength, both for the $L_2$ and $L_\infty$ perturbation norms, is shown in Figure~\ref{fig:ensemble_L2} and Figure~\ref{fig:ensemble_Linfty}. For all perturbation strengths measured by both norms, the ensemble performs better than the individual models.
We see a clear benefit of OOD detector ensembling both on the unperturbed AUROC as well as on the adversarial robustness of the resulting detector. This benefit combines well with the benefit of using the Relative Mahalanobis distance, suggesting that using both could be the correct strategy when deploying OOD detection systems.

\subsection{The effect of image resolution}
\begin{figure}[h!]
    \centering
     \includegraphics[width=0.20\textwidth]{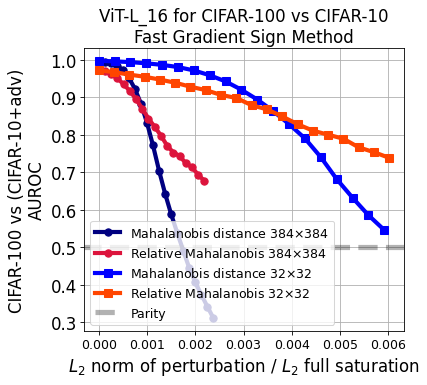}
     \includegraphics[width=0.20\textwidth]{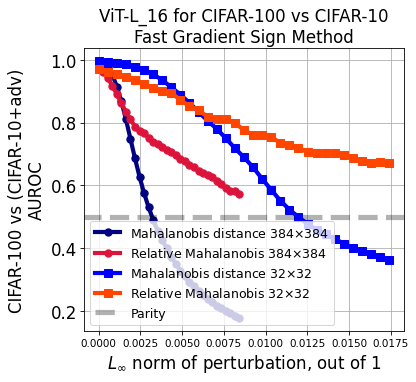}
    \caption{AUROC of CIFAR-100 vs CIFAR-10 where the out-distribution CIFAR-10 is represented by 32 adversarially perturbed images to lower the Mahalanobis distance OOD score (blue) and Relative Mahalanobis distance score (red). We show the perturbation strength measured by its $L_2$ norm (left panel) and $L_\infty$ norm (right panel). The lighter lines show results for images and their gradients at the original CIFAR $32\times32$ resolution, while the darker lines show the $384\times384$ resolution. The lower resolution images are harder to adversarially perturb.}
    \label{fig:resolution}
\end{figure}
\begin{table}[ht]
\begin{center}	
\caption{The effect of image and gradient resolution on OOD robustness. Using differential image upsampling, we show that working with lower resolution images provides adversarial robustness as compared to working with high resolution even for strong near-OOD detectors.}
\vspace{0.5em}
\begin{tabular}{ c|c|c|c|c }

Resolution & \makecell{Post-\\process\\method} & \makecell{AUROC\\before} & \makecell{AUROC\\$\ell_{\infty}$\\1/255} & \makecell{$\Delta$\\AUROC} \\
\hline

32$\times$32 & Maha & 97.98\% & 93.11\% & -4.87\% \\
384$\times$384 & Maha & 97.98\% & 41.33\% & -56.65\% \\

\hline

32$\times$32 & Relative & 97.11\% & 90.13\% & -6.98\% \\
384$\times$384 & Relative & 97.11\% & 71.84\% & -25.27\% \\

\hline

\end{tabular}

\label{tab:resolution}
\end{center}
\end{table}
The input to the Vision Transformer is either $384\times384$ (or $224 \times 224$) while the resolution of both CIFAR-10 and CIFAR-100 is $32\times32$. To resolve that, we upsample images to the correct resolution using the $\mathrm{tf.image.resize}$ function prior to feeding them into the network. This means that the image $\vx$ coming in has the high resolution required, and that the gradient $\vec{g}(\vx) = \partial \mathrm{score} (\vx) / \partial{\vx}$ will be of the same resolution. This gives the attack many more pixels to change and potentially exploit, plausibly leading to an easier to find adversarial example. 

To measure the difference between the adversarial robustness of low and high resolution images, we compared the attacks on the images upsampled prior to their use and gradient computation to working with the low resolutuion images directly. For the latter case, we compute the image score as $\mathrm{score}(\mathrm{resize}(\vx))$ and its derivative as $\partial \mathrm{score}(\mathrm{resize}(\vx)) / \partial \vx$, working directly with the small resolution image and modifying it using the small resolution gradient.

The results for both the standard Mahalanobis distance and the Relative Mahalanobis distance, as well as the perturbation strength $L_2$ and $L_\infty$ norms, are shown in Figure~\ref{fig:resolution} and in Table~\ref{tab:resolution}. The lower resolution images are harder to perturb at a given perturbation strength, however, the benefit (or at least comparable performance at low strength) of the Relative Mahalanobis distance persists.  

\subsection{Exploring far OOD CIFAR-100 vs SVHN}
\begin{figure}[h!]
    \centering
     \includegraphics[width=0.38\textwidth]{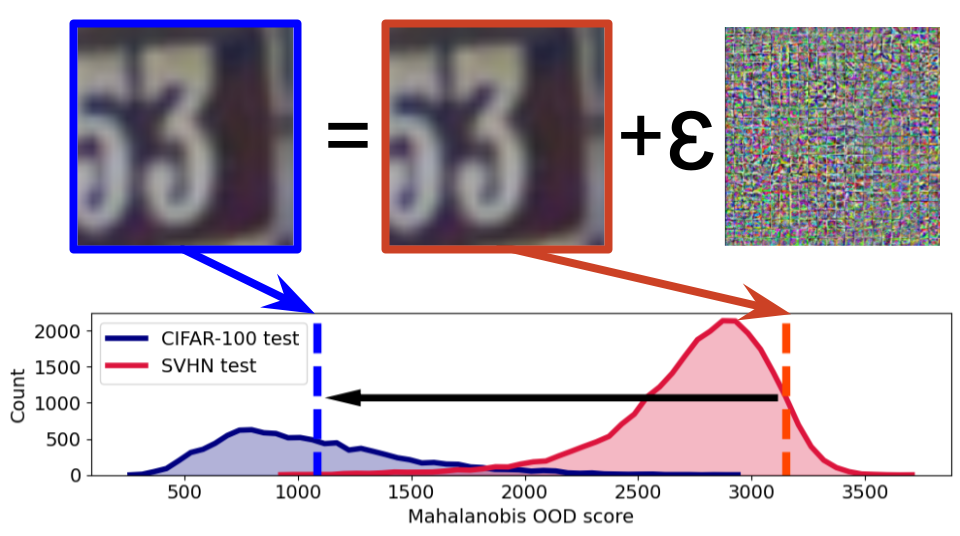}
    \caption{A small adversarial perturbation to the pixels of the out-distribution image (SVHN) changes its out-of-distribution score from $\approx$ 3,000 to a confident in-distribution (CIFAR-100) region at $\approx$ 1,000 even for a state-of-the-art near OOD detection method. $\varepsilon = 10^{-4}$ and the attack used is the Fast Gradient Sign Method applied to the Mahalanobis distance score for a ViT-L$_{16}$ as used in \citep{fort2021exploring}. The unperturbed CIFAR-100 $\to$ SVHN AUROC for this model is 99.40\%.}
    \vspace{-0.5em}
    \label{fig:SVHN_title_figure}
\end{figure}

We studied the adversarial vulnerability on another, easier, far OOD task. In particular, we looked at the CIFAR-100 (in-distribution) vs SVHN (out-distribution) \citep{goodfellow2014multidigit}. We show an example of the adversarial modification in Figure~\ref{fig:SVHN_title_figure}. The very large benefit of the Relative Mahalanobis distance for adversarial robustness of the OOD classification seen for near OOD tasks, such as in Figure~\ref{fig:auroc_L2_ViT-L_16}, Figure~\ref{fig:auroc_Linfty_ViT-L_16} and Table~\ref{tab:main_table}, is not prominent or does not exist at all for this far OOD task. The results are summarized in Table~\ref{tab:SVHN_table}.
\begin{table}[ht]
\begin{center}	
\caption{A comparison of OOD adversarial robustness of the Mahalanobis and Relative Mahalanobis distances for the far OOD CIFAR-100 vs SVHN. }
\vspace{0.5em}
\begin{tabular}{ c|c|c|c } 

	 \makecell{Post-\\process\\method} & \makecell{AUROC\\before} & \makecell{AUROC\\$\ell_{\infty}$\\1/255} & \makecell{$\Delta$\\AUROC} \\
\hline

Maha & 99.40\% & 34.47\% & -64.93\% \\
Relative & 97.19\% & 43.22\% & -53.97\% \\
\hline

\end{tabular}
\vspace{-1.0em}
\label{tab:SVHN_table}
\end{center}
\vspace{-0.35em}
\end{table}

\begin{figure}[h!]
    \centering
     \includegraphics[width=0.21\textwidth]{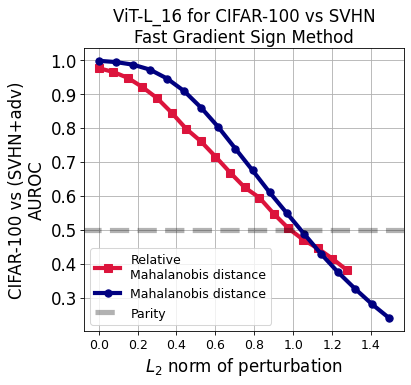}
     \includegraphics[width=0.21\textwidth]{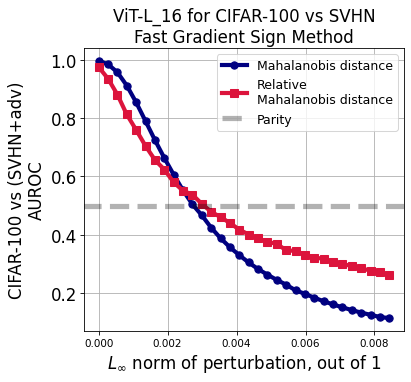}
    \caption{AUROC of CIFAR-100 vs SVHN where the out-distribution SVHN is represented by 128 adversarially perturbed images to lower the Mahalanobis distance (blue) and Relative Mahalanobis distance (red) OOD score. We show the perturbation strength measured by its $L_2$ norm (left panel) and $L_\infty$ norm (right panel). The benefit of the Relative Mahalanobis distance to OOD adversarial robustness is not significant or not as prominted as for the near OOD CIFAR-100 vs CIFAR-10.}
    \vspace{-0.35em}
    \label{fig:cifar100_svhn_pert_size}
\end{figure}

The loss of AUROC from the unperturbed 99.40\% as a function of the $L_2$ and $L_\infty$ norm of the image perturbation are shown in Figure~\ref{fig:cifar100_svhn_pert_size}.
At the $\ell = 1/255$ level of $L_\infty$ perturbation the AUROC is 34.47\%. At the same level with the very same adversary-generation procedure, CIFAR-100 vs CIFAR-10 (near OOD) AUROC drops to 41.33\% (see Table~\ref{tab:main_table} for more details). It seems that, based on this example, there is a weak evidence that far OOD tasks might be more susceptible to adversarial attacks on the OOD score.

\vspace{-0.5em}
\section{Conclusion}
Even very powerful, near out-of-distribution detection methods based on large, pre-trained models, such as the Vision Transformer \citep{fort2021exploring} and multi-modal text-image models, such as CLIP, suffer from severe adversarial vulnerability to their OOD detection score. Well-targeted, small modifications to the image pixels cause these detection systems to change their classification from confidently in-distribution to confidently out-distribution and vice versa. This might come as a surprise given the recent large improvements on near OOD tasks (such as distinguishing CIFAR-100 vs CIFAR-10) these models brought about. We show that orthogonally to their representational robustness that we can infer from their near-OOD performance, they still suffer from a severe \textit{adversarial vulnerability}.

By studying the change in the OOD detectors' AUROC as a function of adversarial perturbation strength, we show that there are easy-to-use and generally applicable approaches to partial remedying this effect: \textbf{ensembling} and the \textbf{Relative Mahalanobis ditance}. The first approach is to ensemble several OOD detectors by averaging their predicted OOD score. The second approach is to use, instead of the standard Maximum of Softmax Probabilities or the more involved Mahalanobis distance post-processing technique, the newly proposed \textit{Relative} Mahalanobis distance \citep{ren2021simple}. We also show that these approaches combine well together.

We hope that by demonstrating this specific non-robustness of even the most powerful approaches to near OOD detection, more research will try to address them. We start off with proposing to use model ensembles and the Relative Mahalanobis distance where possible as an easy to use and cheap fix. However, stronger mitigation techniques will likely have to be employed to meet the frequent safety-critical nature of OOD detection.

\newpage
\clearpage
 \subsection*{Acknowledgements}
 We thank Jie Ren, Huiyi Hu, and Balaji Lakshminarayanan for useful comments and discussions.

\bibliographystyle{unsrt}
\bibliography{main}

\clearpage
\newpage
\appendix

\end{document}